%% file: main.tex
\documentclass[10pt,twocolumn,letterpaper]{article}

\usepackage{cvpr}              % To produce the CAMERA-READY version
\usepackage{graphicx}
\usepackage{amsmath}
\usepackage{amssymb}
\usepackage{booktabs}
\usepackage{array}
\usepackage{multirow}
\usepackage{diagbox}
\newcolumntype{M}[1]{>{\centering\arraybackslash}m{#1}}

\usepackage[normalem]{ulem}

\usepackage[accsupp]{axessibility}  % Improves PDF readability for those with disabilities.

\usepackage[pagebackref,breaklinks,colorlinks]{hyperref}

\usepackage[capitalize]{cleveref}
\crefname{section}{Sec.}{Secs.}
\Crefname{section}{Section}{Sections}
\Crefname{table}{Table}{Tables}
\crefname{table}{Tab.}{Tabs.}

\def\cvprPaperID{5828} % *** Enter the CVPR Paper ID here
\def\confName{CVPR}
\def\confYear{2023}

\makeatletter
\def\thanks#1{\protected@xdef\@thanks{\@thanks
        \protect\footnotetext{#1}}}
\makeatother

\begin{document}

\title{DistilPose: Tokenized Pose Regression with Heatmap Distillation}

\author{Suhang Ye$^{1*}$ \qquad Yingyi Zhang$^{2*}$ \qquad Jie Hu$^{1*}$ \qquad Liujuan Cao$^1$ \\ Shengchuan Zhang$^{1\dagger}$ \qquad Lei Shen$^2$ \qquad Jun Wang$^3$ \qquad Shouhong Ding$^2$ \qquad Rongrong Ji$^1$\\
% $^1$Media Analytics and Computing Laboratory, Department of Artificial Intelligence, School of Informatics, Xiamen University\\
% $^1$Media Analytics and Computing Lab, School of Informatics, Xiamen University, \\
$^1$Key Laboratory of Multimedia Trusted Perception and Efficient Computing, \\Ministry of Education of China, Xiamen University, 
$^2$Tencent Youtu Lab,
$^3$Tencent WeChat Pay Lab33
\thanks{
$^*$~Equal contribution. This work was done when Suhang Ye was an intern at Tencent Youtu Lab.
}
\thanks{
$^{\dagger}$~Corresponding author:~\url{zsc_2016@xmu.edu.cn}
}
}

% \and

\maketitle

%%%%%%%%% ABSTRACT
\begin{abstract}
%Human pose estimation has been dominated by heatmap-based methods due to the superior performance.
%
%Correspondingly, regression-based methods with lower computational cost are more popular on mobile devices.
%

In the field of human pose estimation, 
regression-based methods have been dominated in terms of
speed, while heatmap-based methods are far ahead 
in terms of performance.
How to take advantage of both schemes remains 
a challenging problem.
In this paper, we propose a novel human pose estimation framework
termed DistilPose, which bridges the gaps between 
heatmap-based and regression-based methods.
%
%\textcolor{red}{
Specifically, DistilPose maximizes the transfer of knowledge 
from the teacher model (heatmap-based) 
to the student model (regression-based) 
through Token-distilling Encoder (TDE) and Simulated Heatmaps.
TDE aligns the feature spaces of heatmap-based and 
regression-based models by introducing tokenization, 
while Simulated Heatmaps transfer explicit guidance 
(distribution and confidence) from teacher heatmaps into
student models.
%}
%
Extensive experiments show that the proposed DistilPose can
significantly improve the performance of the regression-based models
while maintaining efficiency. 
Specifically, on the MSCOCO validation dataset, DistilPose-S 
obtains 71.6\% mAP with 5.36M parameters, 2.38 GFLOPs, and 40.2 FPS,
which saves 12.95$\times$, 7.16$\times$ computational cost and is
4.9$\times$ faster than its teacher model with only 0.9 points 
performance drop.
Furthermore, DistilPose-L obtains 74.4\% mAP on MSCOCO validation 
dataset, achieving a new state-of-the-art among 
predominant regression-based models.
Code will be available at \url{https://github.com/yshMars/DistilPose}.
\end{abstract}

%%%%%%%%% BODY TEXT
\section{Introduction}
\label{sec:intro}

2D Human Pose Estimation (HPE) aims to detect the anatomical 
joints of a human in a given image to estimate the poses.
HPE is typically used as a preprocessing module that
participates in many downstream tasks, 
such as activity recognition~\cite{Vats_2022_CVPR,yadav2022arfdnet}, 
human motion analysis~\cite{chen2013survey}, 
motion capture~\cite{moeslund2006survey}, etc.
Previous studies on 2D HPE can be mainly divided into two mainstreams:
\emph{heatmap-based} and \emph{regression-based} methods.
%
%\textcolor{red}{
Regression-based methods have significant advantages in
speed and are well-suited for mobile devices.
However, the insufficient accuracy of regression models
will affect the performance of downstream tasks.
In contrast, heatmap-based methods can explicitly learn
spatial information by estimating likelihood heatmaps,
resulting in high accuracy on HPE tasks.
But the estimation of likelihood heatmaps requires exceptionally high
computational cost, which leads to slow preprocessing operations.
%}
%
%By estimating likelihood heatmaps, heatmap-based models explicitly learn spatially informative features for keypoints, achieving high accuracy and long dominating the research on HPE.
%
%However, the estimation of likelihood heatmaps requires extremely high computational cost, which leads to slow preprocessing operations and severely affects the speed of the downstream tasks.
%
%For this reason, many studies have been proposed to improve the performance of regression-based methods for better efficiency, which incorporate some advanced modules to improve regression performance.
%
%For example, RLE~\cite{RLE} introduced a flow model to gradually increase the constraints on keypoints, while PRTR~\cite{PRTR} and Poseur~\cite{poseur} exploited the self-attention mechanism in the transformer module to capture the relationship between keypoints.
%
%Despite the great success on finding the implicit relationship of keypoints, the performance is still unsatisfactory due to the lack of explicit guidance of heatmaps.
%
%\textcolor{red}{
Thus, how to take advantages of both heatmap-based and 
regression-based methods remains a challenging problem.
%}

\begin{figure}[t]
    \centering
    \includegraphics[width=\columnwidth]
    {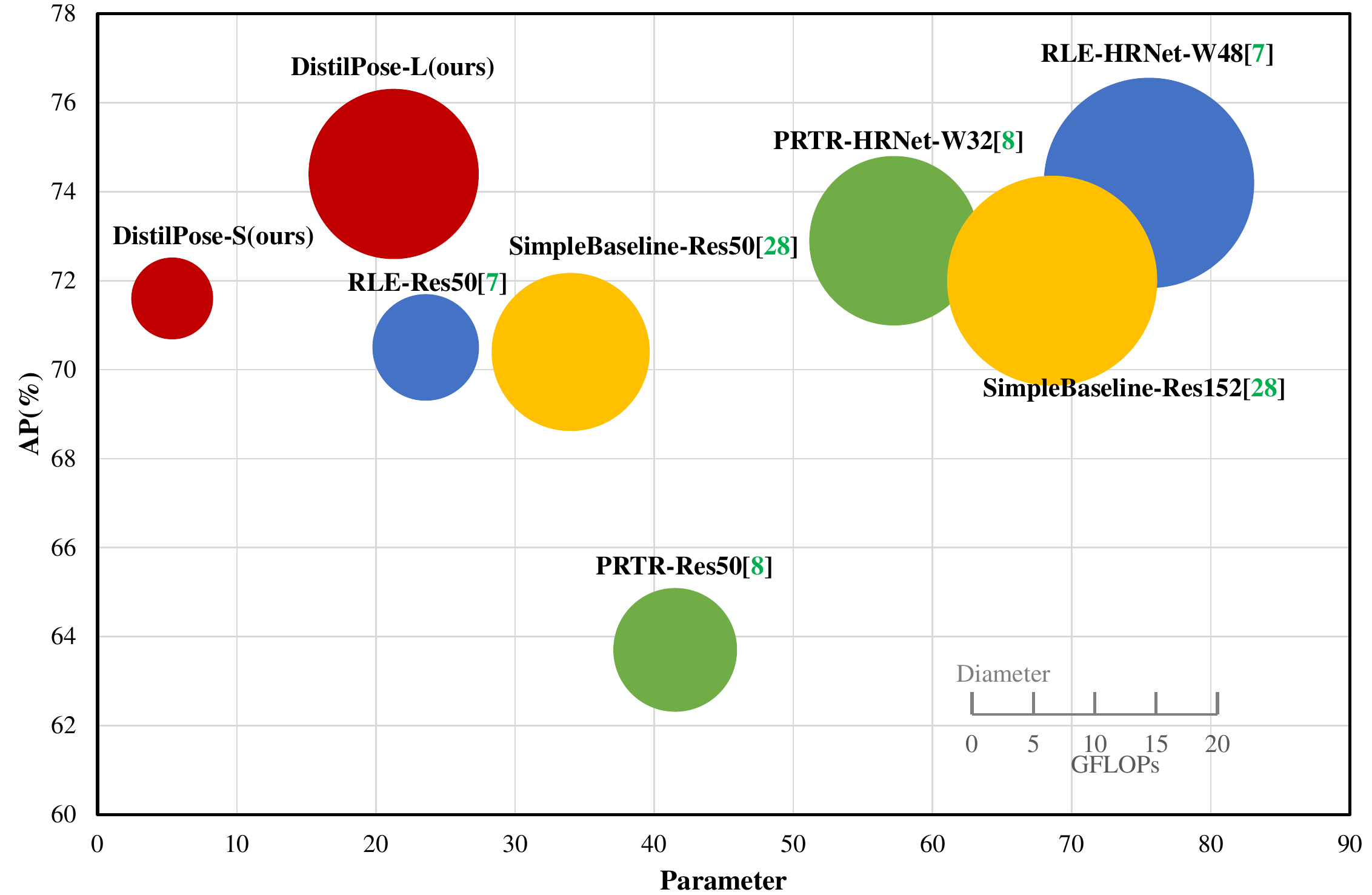}
    \caption{
    \textbf{Comparisons between the SOTA methods and the proposed DistilPose} on MSCOCO \emph{val} dataset.
    Red circles at the upper left corner denote DistilPose.
    DistilPose outperforms SOTA models in terms of accuracy (AP), Parameter and computational cost (GFLOPs). 
    }
    \vspace{-0.6cm}
    \label{fig:Comparison with reg SOTA}
\end{figure}

%Recently, some regression works tried to introduce knowledge from heatmap models to optimize regression performance, such as heatmap pretraining~\cite{RLE}, auxiliary heatmap loss~\cite{tian2019directpose}, etc.
%\textcolor{red}{
One possible way to solve the above problem is to transfer the
knowledge from heatmap-based to regression-based models
~\cite{RLE,tian2019directpose}.
%
%However, due to the following two gaps between heatmap-based and regression-based methods, the aforementioned work can only introduce heatmap knowledge at the backbone level, which brought relatively limited improvement.
However, due to the different output spaces of regression models
and heatmap models 
(the former is a vector, and the latter is a heatmap), 
transferring knowledge between heatmaps and vectors faces the following two problems:
%
%(1) The heatmap-based and regression-based models lack a structure in the head where knowledge can be shared.
(1) The regression head usually vectorizes the feature map 
output by the backbone. 
And much spatial information will be lost through 
Global Average Pooling (GAP) or Flatten operation.
Thus, previous work failed to transfer heatmap knowledge 
to regression models fully.
(2) Compared to the coordinate regression, heatmaps naturally
contain shape, position, and gradient information~\cite{gu2021dive}.
%}
%
Due to the lack of explicit guidance for such information, 
regression-based methods are more difficult to learn the implicit 
relationship between features and keypoints than heatmap-based methods.
In this paper, we propose a novel human pose estimation framework, DistilPose,
%which maximizes the transfer of heatmap-based teacher knowledge to student regression-based models.
which learns to transfer the heatmap-based knowledge from a teacher
model to a regression-based student model.
DistilPose mainly includes the following two components:

%
%(1) A knowledge-transferring module called Token-distilling Encoder (TDE) is designed to transfer the heatmap knowledge in the head level, which consists of a series of transformer encoders.
%\textcolor{red}{
(1) A knowledge-transferring module called Token-distilling 
Encoder (TDE) is designed to align the feature spaces of
heatmap-based and regression-based models 
by introducing tokenization, 
which consists of a series of transformer encoders.
%}
%
TDE can capture the relationship between keypoints 
and feature maps/other keypoints~\cite{TokenPose,transpose}.
%
%(2) We build a Simulated Heatmaps module that mimics heatmaps to get explicit guidance from heatmaps for regression tasks.
%
%Like real heatmaps, Simulated Heatmaps provide 
(2) We propose to simulate heatmaps to obtain the heatmap information 
for regression-based students explicitly.
The resulting Simulated Heatmaps provide two explicit guidelines,
including each keypoint's 2D distribution and confidence.
Note that the proposed Simulated Heatmaps can be inserted 
between any heatmap-based and regression-based methods 
to transfer heatmap knowledge to regression models.
DistilPose achieves comparable 
performance to heatmap-based models with less computational 
cost and surpasses the state-of-the-art (SOTA) regression-based methods.
Specifically, on the MSCOCO validation dataset, DistilPose-S achieves 
$71.6\%$ mAP with $5.36$M parameters, $2.38$ GFLOPs and $40.3$ FPS.
DistilPose-L achieves $74.4\%$ mAP with $21.27$M parameters and 
$10.33$ GFLOPs, which outperforms its heatmap-based teacher model in
performance, parameters and computational cost. 
% 图1
In summary, DistilPose significantly reduces the computation while 
achieving competitive accuracy, bringing advantages from both
heatmap-based and regression-based schemes.
As shown in Figure~\ref{fig:Comparison with reg SOTA}, DistilPose
outperforms previous SOTA regression-based methods, such as
RLE~\cite{RLE} and PRTR~\cite{PRTR} with fewer parameters and GFLOPs.
Our contributions are summarized as follows:
\begin{itemize}
    % \vspace{-0.2cm}
    \item We propose a novel human pose estimation framework, DistilPose, which is the first work to transfer knowledge between heatmap-based and regression-based models losslessly.
    % \vspace{-0.3cm}
    \item We introduce a novel Token-distilling Encoder (TDE) to
    take advantage of both heatmap-based and
    regression-based models.
    With the proposed TDE, the gap between the output space of
    heatmaps and coordinate vectors can be facilitated in a 
    tokenized manner.
    %
    % \vspace{-0.7cm}
    \item We propose Simulated Heatmaps to model explicit heatmap information, including 2D keypoint distributions and keypoint confidences.
    With the aid of Simulated Heatmaps, we can transform the
    regression-based HPE task into a more straightforward learning task
    that fully exploits local information.
    Simulated Heatmaps can be applied to any heatmap-based and
    regression-based models
    for transferring heatmap knowledge to regression models.
    %
    %\item Extensive experiments show that DistilPose significantly boosts the performance of regression-based student models, achieving SOTA performance on both speed and accuracy among regression-based methods.
\end{itemize}

\begin{figure*}[h]
    \centering
    \includegraphics[scale=0.36]{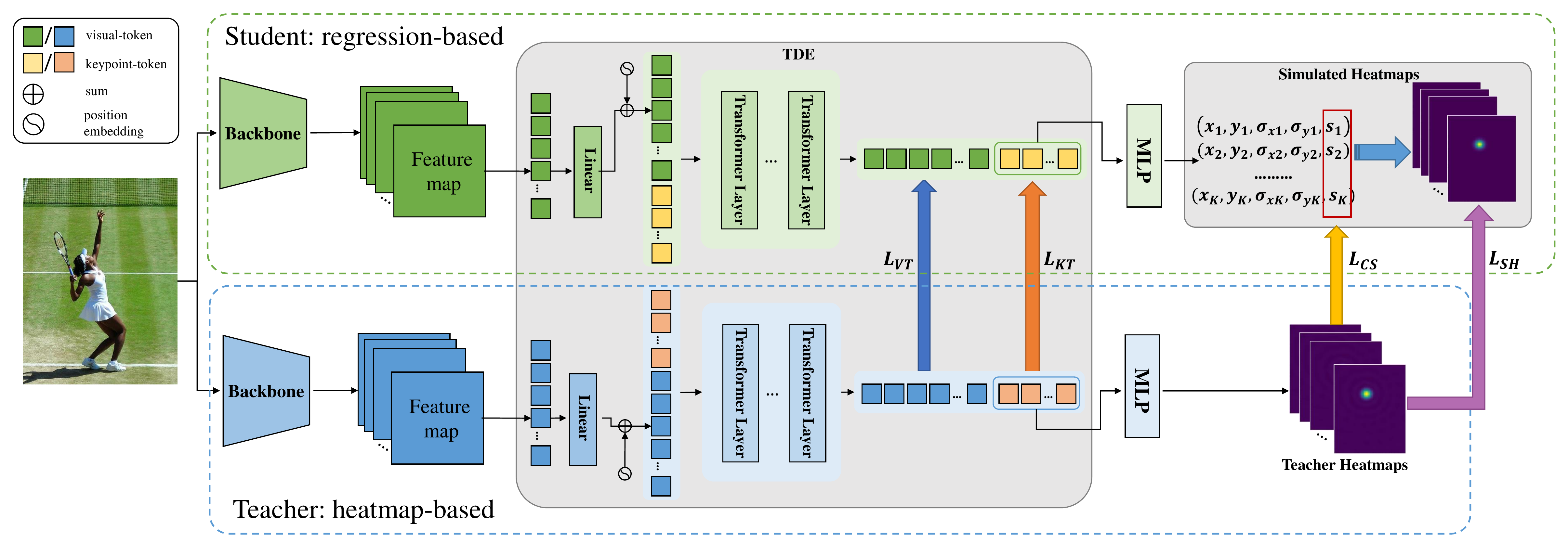}
    \caption{
    Overall architecture of \textbf{DistilPose}.
    During training, a well-trained and fixed heatmap-based teacher 
    provides its knowledge to help the training of regression-based student 
    with TDE and Simulated Heatmaps. 
    %
    %During test phase, only regression-based student is used for inference and no extra test-time is introduced. 
    }
    \vspace{-0.5cm}
    \label{fig:framework}
\end{figure*}

% \vspace{-0.3cm}
\section{Related Work}
\subsection{Heatmap-based \& Regression-based HPE}
%\noindent\textbf{Heatmap-based and Regression-based Pose Estimation.}
%
Heatmap-based pose estimation~\cite{firstHM,hrnet,IntegralHPRegression,simplebaseline,TokenPose,transpose,vitpose, hourglass} 
dominated the area of human pose estimation in terms of performance. 
Some studies~\cite{hrnet, simplebaseline,TokenPose,transpose, hourglass}
constructed novel networks to extract better features.
While others~\cite{DARK, DevilinDetail, papandreou2017towards, tompson2015efficient} 
built upon an optimization perspective trying to mitigate quantization
errors.
In summary, heatmap-based methods made full use of the spatial information
of the feature map and obtain a preferable performance.
However, efficiency is still a certain drawback of heatmap-based methods.

For regression-based methods, Deeppose~\cite{deeppose} is firstly 
proposed to regress the joint coordinates directly.
%
%IEF~\cite{IEF} proposes a process that progressively improves the initial solution by feeding back error predictions. 
%
CenterNet~\cite{centernet} and DirectPose~\cite{tian2019directpose} are
proposed to accomplish multi-person human pose estimation in a
\emph{one-stage} object detection framework, which directly regresses 
the joint coordinates instead of the bounding box.
SPM~\cite{SPM} introduced the root joints to indicate different person
instances and hierarchical rooted human body joints representations to
better predict long-range displacements for some joints. 
Recently, RLE~\cite{RLE} introduced a flow model to capture the underlying
output distribution and gets a satisfying performance.
Although these methods have made great efforts to find the
implicit relationship of keypoints, their performance
improvement is still insufficient due to the lack of explicit
guidance of heatmaps. 

\subsection{Transformer in HPE}
%\noindent\textbf{Transformers for 2D Human Pose Estimation.}
%
Transformer is proposed in \cite{AttentionIsAllYouNeed} and achieves 
great success in Natural Language Processing (NLP). 
Recent studies in vision tasks used Transformer as an alternative 
backbone to CNN for its ability to capture global dependencies.
In the area of 2D human pose estimation, many efforts~\cite{transpose, TokenPose, PRTR, vitpose, PEFormer, poseur, TFPose} 
have been done to incorporate the Transformers.
TFPose~\cite{TFPose} first introduced Transformer to the pose estimation framework in a regression-based manner. 
PRTR~\cite{PRTR} proposed a two-stage and end-to-end regression-based
framework using cascade Transformers and achieves SOTA
performance in regression-based methods. 
TransPose~\cite{transpose} and TokenPose~\cite{TokenPose} introduced
Transformer for heatmap-based human pose estimation achieving 
comparable performance while being more lightweight.
In our work, we introduce the transformer module to assist in
finding potential relationships between keypoints.

\subsection{Distillation in HPE}
%\noindent\textbf{Knowledge Distillation on 2D Human Pose Estimation.}
%
Knowledge Distillation (KD) is formally proposed 
in~\cite{FirstKD_Hinton}, which aims to transfer the teacher's 
learned knowledge to the student model.
In 2D human pose estimation, FPD~\cite{zhang2019fast} 
first used knowledge distillation classically based on the Hourglass 
network.
%
% DKD~\cite{DKD} introduces a light-weight distillator to online 
% distill the knowledge and simplifies body joint localization in 
% video human pose estimation.
%
OKDHP~\cite{onlineKD} introduced an online pose distillation approach
that distills the pose structure knowledge in a one-stage manner. 
ViTPose~\cite{vitpose} also implemented a large-to-small model knowledge
distillation to prove its knowledge transferability.
However, all previous distillation works on human pose estimation 
ignore the knowledge transferring between heatmap-based and 
regression-based methods.
In this work, for the first time, we propose a heatmap-to-regression
distillation framework to take benefits from both schemes. 

\iffalse
\begin{figure*}[t]
    \centering
    \includegraphics[scale=0.6]{figures/backbone_feat_heatmap-3contrib.pdf}
    % }
    \caption{Visualization of backbone feature maps. 
    %
    The three columns in the figure represent direct training,
    heatmap-auxiliary training, and our proposed distillation training,
    respectively.
    %
    It can be seen that under the guidance of heatmap knowledge, the features of the backbone will focus on the human body itself.
    }
    \label{fig:backbone-feat}
\end{figure*}

\begin{figure}[t]
    \centering
    \includegraphics[width=\columnwidth]{figures/attention-map-3contrib.pdf}
    % }
    \caption{Visualization of attention matrix between the keypoint-token
    of ``right-ankle'' (red dot) and visual-tokens. 
    %
    The backbone of these models are all StemNet.
    }
    \label{fig:attention}
\end{figure}
\fi

\begin{figure*}[t]
    \centering
    \setlength{\belowcaptionskip}{-0.5cm}
    \includegraphics[scale=0.43]{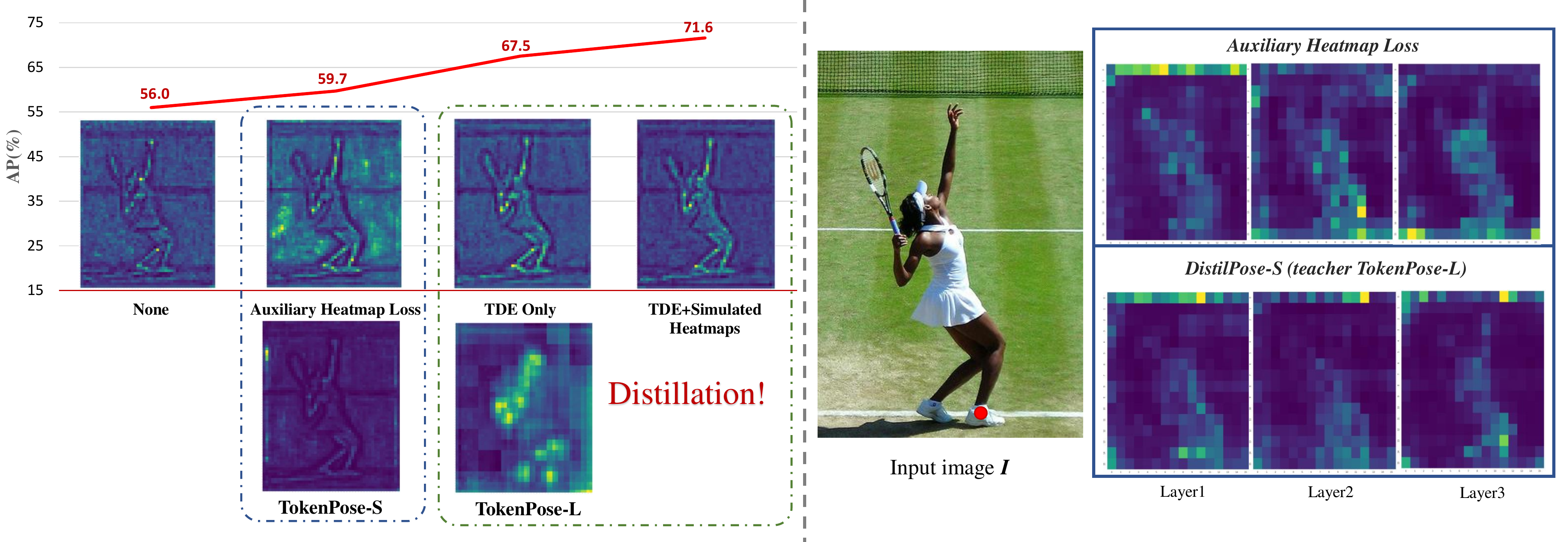}
    \put(-428, -15){(a) Backbone Feature Map}
    \put(-155, -15){(b) Attention Matrix}
    \caption{Visualization of backbone feature maps and attention matrix. 
    (a) The first row represent backbone feature maps
    of direct training, heatmap-auxiliary training, and our proposed
    distillation training (distill TDE only and distill both TDE and Simulated Heatmaps), respectively.
    While the second row represent backbone feature maps
    of the heatmap-based method TokenPose with different
    backbones.
    It can be seen here that the full structure of DistilPose
    is more focused on the human body than the model distill
    TDE only (the former the feature map background is
    \textbf{darker}).
    (b) These are attention matrix between the keypoint-token
    of ``right-ankle'' (red dot) and visual-tokens.
    }
    \label{fig:feat-map}
\end{figure*}

% \vspace{-0.2cm}
\section{Method}
In this section, we propose a distillation-based human pose estimation
framework DistilPose, the overall framework of which is shown in 
Fig.~\ref{fig:framework}.
In our proposed DistilPose, the teacher is a heatmap-based model, and 
the student is a regression-based model.
We transfer the heatmap knowledge of the teacher model to the student model
during training and only use the faster student model in the inference
stage.
DistilPose mainly consists of two modules: Token-distilling 
Encoder and Simulated Heatmaps.

\subsection{Token-distilling Encoder}\label{sec:shared-tokenized}
%
%The performance advantage of heatmap-based methods has long been well known, thanks to the ability of heatmaps to provide direct, explicit, and soft constraints on keypoints.
% 可以补充一点heatmap pretrain的引用
Previous works have tried to introduce the advantages of heatmap models 
in regression-based methods, such as heatmap pre-training~\cite{RLE}, 
auxiliary heatmap loss~\cite{tian2019directpose}, etc.
However, the predict-heads cannot be aligned due to the misalignment of the output space.
That's why these works can only perform knowledge transfer 
on the backbone, which brings models limited performance improvement.
According to Fig.~\ref{fig:feat-map}(a), the 
heatmap-auxiliary model pays too much attention to regions other than 
the human body.
Hence, we propose a Token-distilling Encoder (TDE) to align the output
space of teacher and student by introducing tokenization.
By introducing aligned tokenized features, the heatmap knowledge is transferred to the student model losslessly.
Thus, the student model learns information that is more focused on the
human body itself, as shown in Fig.~\ref{fig:feat-map}(a).
Specifically, for an input image $\textbf{\emph{I}}$, 
we divide it into several patches
according to the size of $pw\times ph$ to form a visual-token.
Next, we add $\emph{K}$ empty nodes as keypoints-token, 
which are concatenated with visual-token and sent to several
transformer encoder layers of TDE.
Inspired by LV-ViT~\cite{LV-ViT}, 
we align the visual-tokens and keypoint-tokens between the student and 
teacher models to obtain the teacher model's refined attention matrix.
As shown in Fig.~\ref{fig:feat-map}(b), the attention matrix 
in TDE can learn the relationship between 
the keypoint-tokens and the visual-tokens of the corresponding
position.
As for performance improvement, TDE enables our student model 
to achieve much higher performance than the heatmap-auxiliary
training (7.8$\%\uparrow$ in Fig.~\ref{fig:feat-map}(a)).

\begin{figure}[t]
    \centering
    \includegraphics[scale=0.3]
    {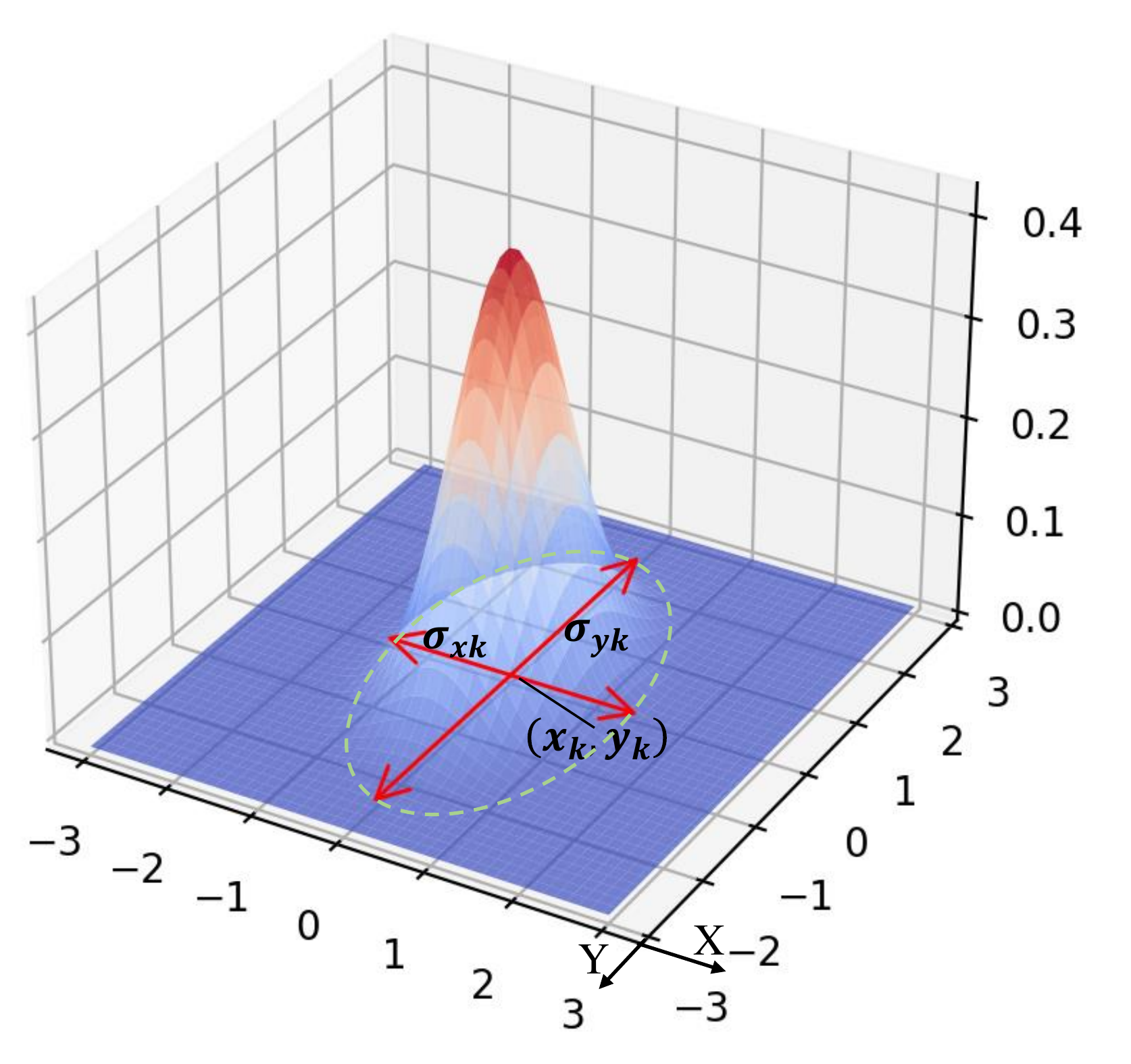}
    \vspace{-0.3cm}
    \caption{Visualization of Basis distribution simulation of 
    $k^{th}$ keypoint in Simulated Heatmaps.
    }
    \vspace{-0.5cm}
    \label{fig:gaussian}
    
\end{figure}

\begin{figure*}[t]
    \centering
    \includegraphics[scale=0.35]{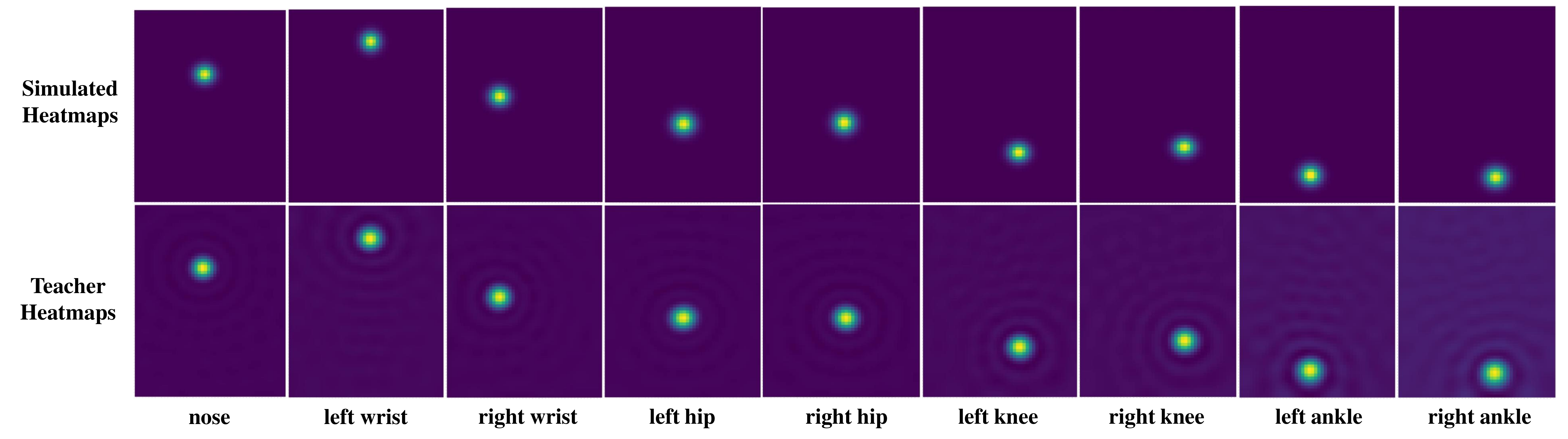}
    % \includegraphics[scale=0.55]{figures/heatmap_more.pdf}
    % }
    \vspace{-0.7cm}
    \caption{Visualization of Simulated Heatmaps and teacher
    heatmaps.
    }
    \vspace{-0.5cm}
    \label{fig:heatmap}
\end{figure*}

% \vspace{-0.2cm}
\subsection{Simulated Heatmaps}
\label{sec:simulated heatmaps module}
%\subsubsection{Basic Distribution Simulation}
%
\textbf{Basic Distribution Simulation}
After the head of the student model obtained the aligned
knowledge with TDE, we began to think about whether there were 
other ways to transfer other knowledge of the teacher model further.
Existing object detection distillation pipelines give us
inspiration~\cite{zheng2022localization}: 
in addition to feature distillation, the output of the 
teacher model can also be used as a soft label to further
distill knowledge to the student model.
Compared with the coordinate vector, 
the distribution with a well-defined shape and good gradient
contained in the heatmap is explicit guide information,
which can prompt the model to pay attention to the local
information around the keypoints.
Not only that, but teacher heatmaps may be closer to reality.
It is because there may be erroneous annotations in ground-truth, 
and the teacher model can filter out these faulty parts 
by summarizing the knowledge during training.
However, heatmaps and coordinate vectors are in different output spaces.
How to introduce the distribution information on the heatmap into the space of the coordinate vector has become an urgent problem to be solved.

For this purpose, we do the opposite and propose the concept of 
Simulated Heatmaps to introduce coordinate vectors into the 
heatmap space.
We let the student model predict both the keypoint coordinates 
and the corresponding $\sigma$ values, using a Gaussian 
distribution to build a virtual heatmap (as shown in Fig.~\ref{fig:gaussian}).
Since the heatmap of the teacher model is not necessarily 
a regular distribution, we predict $\sigma_{x}$ and $\sigma_{y}$ from 
the horizontal/vertical directions, respectively, 
to construct a heatmap closer to the actual one.
The Gaussian distribution is calculated as follows:
%

% \vspace{-0.3cm}
\begin{equation}
  f(x, y) = \frac{1}{2\pi\sigma_{x}\sigma_{y}}e^{-\frac{1}{2}(\frac{(x-\mu_{x})^2}{\sigma_{x}^2}+\frac{(y-\mu_{y})^2}{\sigma_{y}^2})},
  \label{eq:gaussian}
\end{equation}

To align with the ground-truth of the teacher model, 
we ignore the constant coefficient
$\frac{1}{2\pi\sigma_{x}\sigma_{y}}$ 
multiplied by the Gaussian distribution, 
%Therefore, we ignore the constants in the Gaussian distribution,
and the final formula for simulating a heatmap for $k_{th}$ 
keypoint can be summarized as:
%
% \vspace{-0.3cm}
\begin{equation}
  H_{k}(x, y) = e^{-\frac{1}{2}(\frac{(x-\mu_{xk})^2}{\sigma_{xk}^2}+\frac{(y-\mu_{yk})^2}{\sigma_{yk}^2})},
  \label{eq:simulated-heatmap}
\end{equation}

\noindent where $0<x\leq hW$, $0<y\leq hH$ and $0<k\leq \emph{K}$. $hW$ and $hH$
are the width and height of the heatmap, $\emph{K}$ is the total number 
of keypoints, and $H_{k}$ refers to the $k^{th}$ simulated heatmap.
$(\mu_{xk}, \mu_{yk})$ is the predicted keypoints coordinate, 
and $(\sigma_{xk}, \sigma_{yk})$ is the corresponding deviation pair.
In the end, we get $\emph{K}$ heatmap as shown in Fig.~\ref{fig:heatmap}.
%

%\subsubsection{Confidence Distillation}
%
\textbf{Confidence Distillation.} 
In addition to distribution information, 
heatmaps also provide keypoint confidence, 
which helps filter out incorrect model predictions 
and is critical in industrial applications.
However, our simulated heatmap defaults to 1 at the peak, 
which cannot be included in the calculation as a valid confidence.

To this end, we prompt the student model to predict the confidence 
of keypoints $s_k$ directly.
Each keypoint confidence $s_k$ corresponds to the keypoint
coordinate and is constrained by the corresponding value of
the coordinate on the teacher's heatmap.
The keypoint confidence $s_k$ will be accumulated in the human 
instance score, which is calculated as follows:
%
% \vspace{-0.3cm}
\begin{equation}
  s^{human} = s^{bbox}*\frac{\sum_{i=1}^{K} s_{k}}{K},
  % \vspace{-0.3cm}
  \label{eq:bbox-score}
\end{equation}
% \vspace{-0.2cm}
%
\noindent where $s^{human}$ is the prediction score of the human instance, 
and $s^{bbox}$ refers to the confidence score given by the human detector.
%
% \vspace{-0.2cm}
\subsection{Loss Function}
For TDE, we perform distillation for 
visual-tokens and keypoint-tokens, respectively, 
and the corresponding loss functions are:
%
% \vspace{-0.3cm}
\begin{equation}
\begin{aligned}
  L_{KT} = MSE(\textbf{\emph{KT}}_{h}, \textbf{\emph{KT}}_r),\\
  L_{VT} = MSE(\textbf{\emph{VT}}_{h}, \textbf{\emph{VT}}_r),
  \label{eq:dist-loss}
\end{aligned}
% \vspace{-0.2cm}
\end{equation}

\noindent where $\textbf{\emph{KT}}\in\mathbb{R}^{K \times D}$ and
$\textbf{\emph{VT}}\in\mathbb{R}^{P \times D}$ refer to
keypoint-tokens and visual-tokens.
$P$ is the number of patches, and $D$ refers to the dimension 
of tokens.
The subscript \emph{h} and \emph{r} refer to the heatmap-based
and regression-based model accordingly. 
$MSE$ refers to the Mean Squared Error Loss used to measure the
difference between teacher and student
For the Simulated Heatmaps, we try to align the simulated ones to 
the teacher heatmaps, and the corresponding loss function is:
%
% \vspace{-0.3cm}
\begin{equation}
  L_{SH} = \sum_{i=1}^{K}MSE(H_k, H_{Tk}),
  \label{eq:reg2hm-loss}
  % \vspace{-0.2cm}
\end{equation}

\noindent where $H_{Tk}$ indicates the $k^{th}$ teacher heatmap.
For the peak constraint of the Simulated Heatmaps, we design 
the following loss function:
% \vspace{-0.3cm}
\begin{equation}
  L_{CS} = \sum_{i=1}^{K}| H_{Tk}([x_k], [y_k]) - s_k |,
  \label{eq:score-loss}
  % \vspace{-0.2cm}
\end{equation}

\noindent where $(x_k, y_k), k\in[1,K]$ refers to student coordinate predictions. 
$[\cdot]$ indicates the round-off operation.

In summary, we train regression-based student models simultaneously 
under task supervision and knowledge distillation.
The overall loss function of our distillation framework is 
as follows:
% \vspace{-0.3cm}
\begin{equation}
  L = L_{reg} + \alpha_1L_{KT} + \alpha_2L_{VT} + \alpha_3L_{SH} + \alpha_4L_{CS}.
  \label{eq:total-loss}
  % \vspace{-0.2cm}
\end{equation}

\noindent where $L_{reg}$ is the regression task loss for human pose estimation.
We use Smooth L1 Loss as $L_{reg}$ to compute the distance between predicted joints and ground-truth coordinates. 
Invisible keypoints will be filtered, including ground-truth coordinates or predicted coordinates outside the image.
In other word, we did not use these invisible keypoints for loss calculation. 
% \vspace{-0.3cm}
\section{Experiments}

\begin{table*}[t]
\small
\centering
\setlength{\belowcaptionskip}{-0.1cm}
    \scalebox{0.92}{
    \begin{tabular}{|l|c|c|c|c|c|}
    \hline
    \multicolumn{0}{|c|}{\textbf{Methods}} & \textbf{Backbone}  & \textbf{Input Size} & \textbf{Param.(M)} & \textbf{GFLOPs} & \textbf{mAP(\%)} \\
    \hline \hline
    \multicolumn{6}{|l|}{\textit{\textbf{Heatmap-based Methods}}} \\
    \hline   
    % 8-stage Hourglass (ECCV 2016) & Hourglass-8 stacked & 256$\times$192 & 25.1  & 14.3  & 66.9 \\
    % \hline 
    % CPN (CVPR 2018)   & ResNet-50 & 256$\times$192 & 27.0 & 6.2 & 68.6 \\
    % CPN~\cite{CPN}   & ResNet-50 & 384$\times$288 &       &       & 71.6 \\
    SimpleBaselines~\cite{simplebaseline} & ResNet-50 & 256$\times$192 & 34.0 & 8.90 & 70.4 \\
    SimpleBaselines~\cite{simplebaseline} & ResNet-101 & 256$\times$192 & 53.0 & 12.40 & 71.4 \\
    SimpleBaselines~\cite{simplebaseline} & ResNet-152 & 256$\times$192 & 68.6 & 15.70 & 72.0 \\
    \hline
    HRNet~\cite{hrnet} & HRNet-W32 & 256$\times$192 & 28.5 & 7.10 & 74.4 \\
    HRNet~\cite{hrnet} & HRNet-W48 & 256$\times$192 & 63.6 & 14.60 & 75.1 \\
    \hline
    TokenPose~\cite{TokenPose} & stemnet & 256$\times$192 & 6.6 & 2.40 & 72.5 \\
    TokenPose~\cite{TokenPose} & HRNet-W48-stage3 & 256$\times$192 & 27.5 & 11.60 & 75.8 \\
    \hline
    TransPose~\cite{transpose} & ResNet-small & 256$\times$192 & 5.0   & 5.40   & 71.5 \\
    TransPose~\cite{transpose} & HRNet-Small-W48 & 256$\times$192 & 17.5  & 21.80  & 75.8 \\
    \hline \hline
    \multicolumn{6}{|l|}{\textit{\textbf{Distillation-based Methods}}} \\
    \hline 
    OKDHP~\cite{OKDHP} & 2-Stack HG & 256$\times$192 & 13.0  & 25.50  & 72.8 \\
    OKDHP~\cite{OKDHP} & 4-Stack HG & 256$\times$192 & 24.0  & 47.00 & 74.8 \\
    \hline \hline
    \multicolumn{6}{|l|}{\textit{\textbf{Regression-based Methods}}} \\
    \hline 
    % Deeppose (CVPR 2014) & ResNet-50 & 256$\times$192 & 23.58 & 4.04  & 52.6 \\
    % % \hline
    % Deeppose (CVPR 2014) & ResNet-152 & 256$\times$192 & 58.21 & 11.34 & 58.3 \\
    % \hline
    PRTR$^*$~\cite{PRTR} & ResNet-50 & 256$\times$192 & 41.5  & 5.45  & 63.7 \\
    % \hline
    PRTR~\cite{PRTR}   & ResNet-50 & 384$\times$288 & 41.5  & 11.00    & 68.2 \\
    PRTR~\cite{PRTR}  & ResNet-50 & 512$\times$384 & 41.5  & 18.80  & 71.0 \\
    PRTR$^*$~\cite{PRTR}  & HRNet-W32 & 256$\times$192 & 57.2 & 10.23 & 72.9 \\
    PRTR~\cite{PRTR} & HRNet-W32 & 384$\times$288 & 57.2 & 21.60  & 73.1 \\
    PRTR~\cite{PRTR}   & HRNet-W32 & 512$\times$384 & 57.2 & 37.80  & 73.3 \\
    \hline
    RLE~\cite{RLE}  & ResNet-50 & 256$\times$192 & 23.6 & 4.04  & 70.5 \\
    RLE$^*$~\cite{RLE}   & HRNet-W48 & 256$\times$192 & 75.6 & 15.76 & 74.2 \\
    \hline
    % PEFormer-Deit-dino-p8 (PRAI 2022) & ResNet-50 & 256$\times$192 & 36.4 & -  & 70.6 \\
    % PEFormer-Xcit-p16 (PRAI 2022) & ResNet-50 & 384$\times$288 & 40.6 & -  & 70.2 \\
    % PEFormer-Xcit-dino-p16 (PRAI 2022) & ResNet-50 & 384$\times$288 & 40.6 & -  & 70.7 \\
    % PEFormer-Xcit-dino-p8 (PRAI 2022) & ResNet-50 & 256$\times$192 & 40.5 & -  & 71.6 \\
    % PEFormer-Xcit-dino-p8 (PRAI 2022) & ResNet-50 & 384$\times$288 & 40.5 & -  & 72.6 \\
    Poseur~\cite{poseur}  & MobileNetV2 & 256$\times$192 & 11.36 & 0.5  & 71.9 \\
    Poseur~\cite{poseur}   & ResNet-50 & 256$\times$192 & 33.26 & 4.6 & 75.4 \\
    \hline 
    \textbf{DistilPose-S ($Ours$)}   & stemnet & 256$\times$192 & \textbf{5.4} & \textbf{2.38}  & \textbf{71.6} \\
    % \hline
    % 之前的实验结果，现在应该会比这个结果更好
    \textbf{DistilPose-L ($Ours$)}   & HRNet-W48-stage3 & 256$\times$192 & \textbf{21.3} & \textbf{10.33} & \textbf{74.4} \\
    \hline
    \end{tabular}
    }
\vspace{-0.2cm}
\caption{Comparison on MSCOCO \emph{\textbf{val}} dataset. 
    Flip test is used on all methods.
    $^*$ indicates that we re-train and evaluate the models.  
    }
    \vspace{-0.2cm}
\label{COCO val Performance}
\end{table*}

\iffalse
\begin{table}[t]
    \centering
    \small
    %\resizebox{\columnwidth}{!}{
    \begin{tabular}{|c c c c|}
        \hline
         \textbf{Layer} & \textbf{Embedding} & \textbf{Heads} & \textbf{Patch Size}
        \\
        \hline
        12 &192 &8 &4$\times$3  \\
         
         \hline
         
    \end{tabular}%}
    \caption{Architecture configurations of TDE in our framework DistilPose.
    }
    \label{tab:Architecture configurations}
\end{table}
\fi

\begin{table*}
\centering
\small
\setlength{\belowcaptionskip}{-0.5cm}
    \scalebox{0.92}{
    \begin{tabular}{|l|c|c|c|c|c|c|c|c|}
    \hline
    \multicolumn{0}{|c|}{\textbf{Methods}} & \textbf{Backbone}  & \textbf{Input Size} & \textbf{AP(\%)} & \textbf{AP$_{50}$(\%)} & \textbf{AP$_{75}$(\%)} & \textbf{AP$_{M}$(\%)} & \textbf{AP$_{L}$(\%)} \\
    \hline \hline
    % \multicolumn{8}{|l|}{\textit{\textbf{Regression-based Methods}}} \\
    % \hline \hline
    % Deeppose (CVPR 2014) & ResNet-50 & 256$\times$192  & - \\
    % % \hline
    % Deeppose (CVPR 2014) & ResNet-152 & 256$\times$192 & - \\
    % \hline
    % PRTR~\cite{PRTR}  & ResNet-101 & 256$\times$192  & - \\
    % \hline
    PRTR~\cite{PRTR}   & ResNet-101 & 384$\times$288 & 68.8 & 89.9 & 76.9 & 64.7 & 75.8\\
    % \hline
    PRTR~\cite{PRTR}   & ResNet-101 & 512$\times$384 & 70.6 & 90.3 & 78.5 & 66.2 & \textbf{77.7}\\
    \hline
    RLE$^*$~\cite{RLE}   & ResNet-50 & 256$\times$192  & 69.8 & 90.1 & 77.5 & 67.2 & 74.3\\
    \hline
    \textbf{DistilPose-S ($Ours$)}   & stemnet & 256$\times$192 & \textbf{71.0} & \textbf{91.0} & \textbf{78.9} & \textbf{67.5} & 76.8\\
    \hline \hline
    PRTR~\cite{PRTR}   & HRNet-W32 & 384$\times$288 & 71.7 & 90.6 & 79.6 & 67.6 & 78.4\\
    % \hline
    PRTR~\cite{PRTR}   & HRNet-W32 & 512$\times$384 & 72.1 & 90.4 & 79.6 & 68.1 & 79.0\\
    \hline    
    RLE$^*$~\cite{RLE}   & HRNet-W48 & 256$\times$192 & \textbf{73.7} & 91.4 & \textbf{81.4} & \textbf{71.1} & 78.6\\
    \hline 
    \textbf{DistilPose-L ($Ours$)}   & HRNet-W48-stage3 & 256$\times$192 & \textbf{73.7} & \textbf{91.6} & 81.1 & 70.2 & \textbf{79.6}\\
    \hline
    \end{tabular} 
    } 
    
\caption{Comparison on MSCOCO \emph{\textbf{test-dev}} dataset.
$^*$ indicates that we re-train and evaluate the models.  }
\label{COCO test-dev Performance}
% \vspace{-0.3cm}
\end{table*}

\begin{table*}[th]
    \centering
    \small
    \setlength{\belowcaptionskip}{-0.5cm}
    \scalebox{0.92}{
    \begin{tabular}{|c|c|c|c|c|c c c c|}
        \hline
        \textbf{Model} & \textbf{Role} & \textbf{Backbone} & \textbf{Methods} & \textbf{Ex-post.} & \textbf{AP(\%)} &
        \textbf{Param(M)} & \textbf{GFLOPs} & \textbf{FPS}
        \\
        \hline \hline
        % Poseur~\cite{poseur} & SOTA & MobileNetv2 & regression & - & 71.9 & 11.36 & 0.49 & 8.5 \\
        % \hline
        % Poseur~\cite{poseur} & SOTA & ResNet-50 & regression & - & 75.4 & 33.26 & 4.6 & 8.2 \\
        %  \hline
        % Poseur~\cite{poseur} & SOTA & HRNet-W32 & regression & - & 76.9 & 38.19 & 7.4 & 4.4 \\
        % \hline
        % Poseur~\cite{poseur}$^\dagger$ & SOTA & HRNet-W48 & regression & - & 78.8 & 74.27 & 33.6 & 4.3 \\
        %  \hline \hline
         TokenPose$^*$ & Teacher & HRNet-W48 & heatmap & Y & 75.2 & 69.41 & 17.03 & 7.8 \\
        \hline
         TokenPose$^*$ & Teacher & HRNet-W48 & heatmap & N & 72.5 & 69.41 & 17.03 & 8.2 \\
         \hline
         DistilPose-S & Student & stemnet & regression & - & 71.6 (0.9$\downarrow$) & 5.36 (12.95$\times \downarrow$) & 2.38 (7.16$\times\downarrow$) & 40.2 (4.90$\times\uparrow$) \\
         \hline
         DistilPose-L & Student & HRNet-W48-s3 & regression & - & 74.4 (1.9$\uparrow$) & 21.27 (3.26$\times\downarrow$) & 10.33 (1.65$\times\downarrow$) & 13.7 (1.72$\times\uparrow$) \\
         \hline
    \end{tabular}
    }
    \caption{
    Comparison with Teacher Model.
    \textbf{Ex-post.} = extra post-processing, which means extra post-processing used to refine the heatmap-to-coordinate transformation during inference.
    We compute the multiples in comparison with non extra post-processing heatmap-based method. 
    $^*$ indicates that we re-train and evaluate the models. 
    %
    % \textcolor{blue}{$^\dagger$ indicates that the Poseur model uses a higher resolution~(384$\times$288), instead of default resolution~(256$\times$192).}
    %
    And HRNet-W48-s3 is short for HRNet-W48-stage3.
    }
    \label{tab:fps}
    %\vspace{-0.25cm}
\end{table*}

In this section, we evaluate the proposed distillation framework on the MSCOCO dataset. What's more, we carry out a series of ablation studies to prove the effectiveness and validity of DistilPose. 
The implementation of our method is based on MMPose~\cite{mmpose2020}.
%

% \vspace{-0.2cm}
\subsection{Implementation Details}
% \vspace{-0.1cm}
\subsubsection{Datasets}
% \vspace{-0.1cm}
%
We mainly conduct our experiments on the MSCOCO dataset~\cite{coco}.
The MSCOCO dataset contains over 200k images and 250k human instances. 
Each human instance is labeled with $\emph{K}$ = 17 keypoints 
representing a human pose. 
Our models are trained on MSCOCO train2017 with 57k images and evaluated 
on both MSCOCO val2017 and test-dev2017, which contain 5k and 20k 
images, respectively.
Furthermore, the ablation experiments are conducted on the MSCOCO
\emph{val} dataset.
We mainly report the commonly used standard evaluation metric Average
Precision(AP) as previous works done on the MSCOCO dataset.
%
% \vspace{-0.5cm}
\subsubsection{Training} 
% \vspace{-0.2cm}
We follow the top-down human pose estimation paradigm. 
All input images are resized into 256$\times$192 resolution. 
We adopt a commonly used person detector provided by SimpleBaselines~\cite{simplebaseline} with 56.4\% AP for the MSCOCO \emph{val} dataset and 60.9\% AP for the MSCOCO \emph{test-dev} dataset.
All the models are trained with a batch size of 64 images per GPU, 
using 8 Tesla V100 GPUs.
We adopt Adam as our optimizer and train the models for 300 epochs. 
The base learning rate is set to 1e-3, and decays to 1e-4 and 1e-5 at the 200$^{th}$ and 260$^{th}$ epoch, respectively.
We follow the data augmentation setting in HRNet~\cite{hrnet}.
We empirically set the hyper-parameters to 
$\alpha_1=\alpha_2=5e-4, \alpha_3=1, \alpha_4=1e-2$.
%
% \vspace{-0.4cm}
\subsubsection{Model Setting} 
If not specified, the teacher model we use is a 
heatmap-based model with a performance of 
72.5\% (75.2\% if extra-post-processing~\cite{DARK} is used 
during inference) at the cost of 69.41M parameters and 17.03 GFLOPs, 
which adopts HRNet-W48 as its backbone.
DistilPose-S adopts a lightweight backbone named \emph{stemnet} 
from TokenPose~\cite{TokenPose}, 
which is widely used to downsample the feature map 
into $1/4$ input resolution quickly.
Stemnet is a very shallow convolutional structure and is 
trained from scratch. 
Besides, DistilPose-L uses feature map output by 
HRNet-W48~\cite{hrnet} at the 3$^{rd}$ stage, 
following the same setting as
TokenPose-L~\cite{TokenPose}.
For the architecture configurations of TDE of DistilPose, the num of transformer layers is 12, embedding dim is 192, the num of heads is 8, and patch size $p_w\times p_h$ is 4$\times$3.

% % 需要展示的实验
% 1. COCO VAL上的性能  [done]
% 2. COCO test上的性能（之后跑） [done]
% 3. 与teacher对比的性能， [done]
% <消融实验>
% 1. 各部分loss的逐步增益递进,体现蒸馏前后的一个性能对比  [done]
% 2. distribution guiding:固定的sigma/一维sigma/二维sigma 
% 3. 在CNN上的性能增益：simplebaseline到deeppose [done]
% 4. 使用越强的teacher，性能越好。证明一个我们的性能 [done]
% <Visualization>
% 1. attention matrix的可视化对比，对比蒸馏前后的attention matrix。
% 2. sigma的可视化
% 3. teacher heatmap比高斯分布生成的heatmap更贴近真实分布 [optional]
% 4. score分支的有效性证明 [optional]

% \begin{table}[t]
%     \centering
%     \small
%     \setlength{\belowcaptionskip}{-0.5cm}
%     \resizebox{\columnwidth}{!}{
%     % \scalebox{0.92}{
%     \begin{tabular}{|c|c|c|c|c|c|c|}
%         \hline
%         Method & Backbone & Resolution & Param & GFLOPs & mAP & FPS \\
%         \hline
%         \multirow{4}{*}{Poseur} & MobileNetV2 & 256$\times$192 & 11.36M & 0.5 & 71.9\% & 12.1 \\
%         % \hline
%          & ResNet-50 & 256$\times$192 & 33.26M & 4.6 & 75.4\% & 12 \\
%         % \hline
%          & HRNet-W32 & 256$\times$192 & 38.19M & 7.4 & 76.9\% & 5.5 \\
%         %\hline
%          & HRNet-W48 & 384$\times$288 & 74.27M & 33.6 & 78.8\% & 5.4\\
%         \hline
%         \multirow{2}{*}{DistilPose} & stemnet & 256$\times$192 & 5.36M & 2.4 & 71.6\% & 40.2 \\
%          & HRNet-W48-s3 & 256$\times$192 & 21.27M & 10.3 & 74.4\% & 13.7 \\
%         \hline
%     \end{tabular} 
%     }
%     \caption{Comparison between Poseur and DistilPose on MSCOCO \emph{\textbf{val}} dataset.}
%     \label{tab:poseur}
% \end{table}

\subsection{Main Results}
% \vspace{-0.2cm}
\subsubsection{Comparison with SOTA Regression Methods}
% \vspace{-0.2cm}
%
We compare the proposed DistilPose with the
SOTA regression-based methods on MSCOCO \emph{val} 
and \emph{test-dev} dataset, 
and the experimental results are shown in 
Table~\ref{COCO val Performance} and Table~\ref{COCO test-dev Performance},
respectively.
Also, a line chart is drawn for more visual comparison in
Figure~\ref{fig:Comparison with reg SOTA}.
We mainly compare our methods with PRTR~\cite{PRTR} and RLE~\cite{RLE}.
Since PRTR is also Transformer-based methods that
achieved SOTA performance among regression-based methods, 
RLE still dominates regression-based methods. 
We further compare DistilPose with the latest regression-based SOTA method Poseur~\cite{poseur} in our supplementary materials. 

Specifically, DistilPose-S achieves 71.6\% with 5.36M and 2.38  GFLOPs,
which outperforms PRTR-Res50 by 7.9\% at the same input resolution
256$\times$192 while reducing 36.14M (87.1\%$\downarrow$) parameters 
and 3.07 GFLOPs (56.3\%$\downarrow$).
Even if PRTR adopts larger input resolutions (384$\times$288,
512$\times$384), DistilPose still performs better using 
256$\times$192 input resolution. 
%
% \sout{
% For the recent SOTA method Poseur, its FPS is only 0.63$\times$ of 
% our largest model DistilPose-L (as shown in Table~\ref{tab:fps}), 
% although it only uses a lightweight backbone MobileNetv2 
% with fewer parameters and GFLOPs.
% %
% This proves that our DistilPose-L also outperforms Poseur (2.7\%$\uparrow$) on the premise that it is faster.
% }
%
% The recent SOTA method Poseur~\cite{poseur} has FPN feature and well-designed encoder/decoder, which brings a larger computational burden while achieving higher performance.
% %
% Thus, as shown in Tabel~\ref{tab:poseur}, 1) when with the same backbone~(HRNet-W48), DistilPose achieves much higher FPS while Poseur achieves higher accuracy.
% %
% 2) When with similar AP (DistilPose-S (71.6\%) \& Poseur-MobileNetV2 (71.9\%)),
% %
% the FPS of DistilPose-S is 4.73 times that of Poseur.

%

%
Compared to non-Transformer SOTA algorithms, DistilPose exceeds RLE in
performance, parameter, and computation simultaneously on MSCOCO \emph{val}
dataset.
As shown in Table~\ref{COCO test-dev Performance}, 
we can see that DistilPose performs much better than RLE on 
large human instances.
Since RLE is the first regression-based work with superior 
performance to contemporary heatmap-based methods and 
DistilPose outperforms RLE, we also claim that DistilPose achieves
comparable performance on par with heatmap-based methods, 
as shown in Table~\ref{COCO val Performance}.
%

\iffalse
%

Besides, as can be seen from Table~\ref{COCO val Performance}, the performance of our proposed DistilPose reaches a level comparable to heatmap-based methods, on the premise of keeping the number of parameters and GFLOPs much lower.

%
\fi
% \vspace{-0.3cm}
\subsubsection{Comparison with Teacher Model}
We conduct experiments to compare the performance of the student and 
teacher models in the dimension of AP, Parameter, GFLOPs, and FPS on 
the MSCOCO \emph{val} dataset. 
Extra post-processing~\cite{DevilinDetail, DARK} is always used in the heatmap-based model for the heatmap-to-coordinate transformation during inference to eliminate quantization error, significantly improving performance, but is also followed by extra test-time overhead. 
In this part, we remove the extra post-processing~\cite{DARK} for more fairly exploring the advantages and disadvantages of heatmap-based and regression-based methods. 
We report the results between student and teacher without extra
post-processing.
As shown in Table~\ref{tab:fps}, DistilPose-S sacrifices 0.9\% precision but dramatically improves in the reduction of parameter, computation, and test-time overhead.
DistilPose-L comprehensively outperformed the heatmap-based teacher. 
That's because DistilPose is not only immune to quantization error of heatmaps,
but also maintains the structural advantages of the regression model~\cite{PIPNet}.
Even if compared to the heatmap-based teacher with extra post-processing, DistilPose-L still achieves comparable performance with much less resource consumption.
The experiments above demonstrate that DistilPose can aggregate heatmap and coordinate information and benefit from both schemes.

% \vspace{-0.2cm}
\subsection{Ablation Study}
\iffalse
%
We perform a series of ablation studies on the MSCOCO \emph{val} dataset to demonstrate the effectiveness of DistilPose.
%
The results presented are based on DistilPose-S, with an input resolution of 256$\times$192.
%
\fi
% \vspace{-0.2cm}
\subsubsection{Performance Gain from Different Parts}
% \vspace{-0.1cm}
% 
In this subsection, we conduct several ablation experiments to show how each type of knowledge transfer helps the training of the regression-based student.
As shown in Table~\ref{tab:EachTypeKD}, all types of proposed knowledge transfer benefit the regression-based model. 
From the perspective of knowledge levels, TDE and
Simulated Heatmaps each bring an improvement of 
11.5\% and 8.1\%, respectively.
In the simulation heatmap module, the Gaussian simulation heatmaps
combined with the confidence prediction can improve the
contribution of the two by 0.5\%.
This proves that confidence predictions and Simulated Heatmaps
have mutually reinforcing effects within the model.
The combination of all proposed knowledge transfers brings 
the best performance, 
which significantly improves the performance by 15.6\% 
comparing to the non-distillation regression-based student model. 

\begin{table}[t]
    \centering
    \small
    \scalebox{0.92}{
    \begin{tabular}{|c|p{0.6cm}<{\centering} p{0.6cm}<{\centering}|c c|c c|}
        \hline
         \multirow{2}{*}{\textbf{Distillation}} & \multicolumn{2}{p{0.6cm}<{\centering}}{\textbf{Simulated Heatmaps}} &
         \multicolumn{2}{|c|}{\textbf{TDE}} & \multirow{2}{*}{\textbf{AP}} & \multirow{2}{*}{\textbf{Improv.}} \\
         \cline{2-5}
         ~ & $L_{CS}$ & $L_{SH}$ & $L_{KT}$ & $L_{VT}$ &~&~ \\
         \hline \hline
         No & - & - & - & - & 56.0\% & -  \\
         \hline
         \multirow{7}{*}{Yes} & \checkmark & & & & 63.2\% & +7.2\%  \\
         ~ &  & \checkmark & ~ & ~ & 56.4\% & +0.4\% \\
         ~ & \checkmark & \checkmark & ~ & ~ &  64.1\% & +8.1\%  \\
         \cline{2-7}
         ~ &  &  & \checkmark &  &  67.1\% & +11.1\% \\
         ~ &  &  &  & \checkmark &  61.7\% & +5.7\% \\
         ~ &  &  & \checkmark & \checkmark &  67.5\% & +11.5\% \\
         \cline{2-7}
         ~ &\checkmark & \checkmark & \checkmark & \checkmark &  71.6\% & +15.6\% \\
         \hline
    \end{tabular} }
    \caption{Ablation studies for different types of knowledge distillation.
    All types of proposed knowledge transferring benefit the regression-based model, and the combination of all proposed knowledge transferring brings the best performance.
    \textbf{Improv.} = Improvement.
    }
    \vspace{-0.8cm}
    \label{tab:EachTypeKD}
\end{table}

% \vspace{-0.3cm}
\subsubsection{Better Teacher, Better Performance}
% \vspace{-0.2cm}
%
We conduct experiments to study the performance gain between 
different students and teachers.
As shown in Table~\ref{tab:betterTeacher}, we demonstrate that 
the student models will perform better if they are guided by 
stronger teacher models. 
This suggests that the performance of DistilPose we report is not 
the upper-bound accuracy of our distillation framework 
but an example to illustrate validity. 
In practice, we could get help from previous algorithms to train 
a more potent teacher, which can be further used to enhance 
the performance of the student model. 
Or we can simply expand the teacher model's capacity, 
finding a trade-off between performance and training memory limitation 
to maximize the utilization of training resources 
and not introduce any test-time overhead during inference. 
%

%这边要不要跑一个5分支的去说，然后消除多分支正则化的影响
\begin{table}[t]
    \centering
    \small
    \scalebox{0.92}{
    \begin{tabular}{|c|c|c|c|}
        \hline
        \diagbox{\textbf{Student}}{\textbf{Teacher}} & \textbf{None} & \textbf{stemnet} & \textbf{HRNet-W48} \\
         \hline \hline
         stemnet & 56.0\% & 63.6\% & 71.6\% \\
         \hline
         HRNet-W48-stage3 & 63.0\% & 66.8\% & 74.4\% \\
         \hline
    \end{tabular} 
    }
    \vspace{-0.2cm}
    \caption{Ablation studies on different volumes of student and teacher,
    which demonstrates that regression-based student model performs 
    better with a stronger teacher.
    }
    \vspace{-0.4cm}
    \label{tab:betterTeacher}
\end{table}

% \vspace{-0.5cm}
\subsubsection{Generalization on CNN-based Model}
% \vspace{-0.2cm}
%
We also conduct an ablation study on a CNN-based model to show its generalizability.
We adopt Deeppose~\cite{deeppose} as the student and
SimpleBaselines~\cite{simplebaseline} as the teacher.
Both are the most concise algorithms in regression-based and heatmap-based models, respectively. 
We introduce Simulated Heatmaps to transfer heatmap-level knowledge as we do in DistilPose, and ignore TDE for the CNN-based model. 
Both student and teacher adopt ResNet-50 as the backbone, and the experimental results in Table~\ref{tab:on CNN} show that Simulated Heatmaps adapted from DistilPose improve the regression-based student's performance significantly with no extra test-time overhead.

\begin{table}[t]
    \centering
    \small
    \setlength{\belowcaptionskip}{-0.3cm}
    \scalebox{0.92}{
    \begin{tabular}{|c|c c|c c|}
        \hline
        \textbf{Model} &
        \begin{tabular}{c}
             \textbf{Simulated}\\
             \textbf{Heatmaps}
        \end{tabular} 
        % \textbf{$L_{HGL}$}
        & \textbf{Role} & \textbf{mAP} & \textbf{Improv.} \\
         \hline \hline
         SimpleBaseline & - & Teacher & 70.4\% & - \\
         \hline
         Deeppose & $\times$ & - & 52.6\% & - \\
         \hline
         Deeppose & \checkmark & Student & 59.7\% & + 6.9\% \\
         \hline
    \end{tabular} }
    \vspace{-0.2cm}
    \caption{Generalization on CNN-based Model.
    We only implement Simulated Heatmaps to transfer heatmap-level knowledge 
    on CNN-based models and get a significant performance improvement. 
    \textbf{Improv.} = Improvement.
    }
    \vspace{-0.2cm}
    \label{tab:on CNN}
\end{table}

% \begin{table}[t]
%     \centering
%     \small
%     % \setlength{\abovecaptionskip}{-0.1cm}
%     % \setlength{\belowcaptionskip}{-0.3cm}
%     \scalebox{0.8}{
%     \begin{tabular}{|c|c|c|c|c|c|}
%     \hline
%          & mAP & Shoulder(l) & Shoulder(r) & Knee(l) & Knee(r)\\
%         \hline 
%        $\sigma$  & 71.2\% & 2.09 & 2.12 & 2.09 & 2.14 \\
%        $(\sigma_x, \sigma_y)$ & 71.6\% & (2.10, 2.06) & (2.04, 2.08) & (2.05, 1.89) & (2.08, 1.86) \\
%         \hline
%     \end{tabular} 
%     }
%     \caption{Comparison between single deviation and horizontal/vertical deviations for Basic Distribution Simulation.}
%     \label{tab:comparison_deviation}
% \end{table}

% % \vspace{-0.3cm}
% \subsubsection{Deviation for Basic Distribution Simulation}
% % \vspace{-0.1cm}
% % DistilPose predicts two deviations $\sigma_x$ and $\sigma_y$ from horizontal/vertical directions respectively to better construct a heatmap using Gaussian distribution, as illustrated in Sec.\ref{sec:simulated heatmaps module}.
% %
% We conduct an ablation study to demonstrate that predicting deviations from different directions~($\sigma_x$/$\sigma_y$ for horizontally/vertical respectively) can better help student model learn the distribution information from teacher heatmaps than predicting one deviation $\sigma$ for all directions, as shown in Tab.\ref{tab:comparison_deviation}.
% %
% We also provide a set of sample comparisons of predicted deviations on the same image in Tab.\ref{tab:comparison_deviation}.
% %
% The visualization of simulated heatmaps is shown in Fig\ref{fig:heatmap}.

\begin{figure}[t]
    \centering
    \includegraphics[width=\columnwidth]{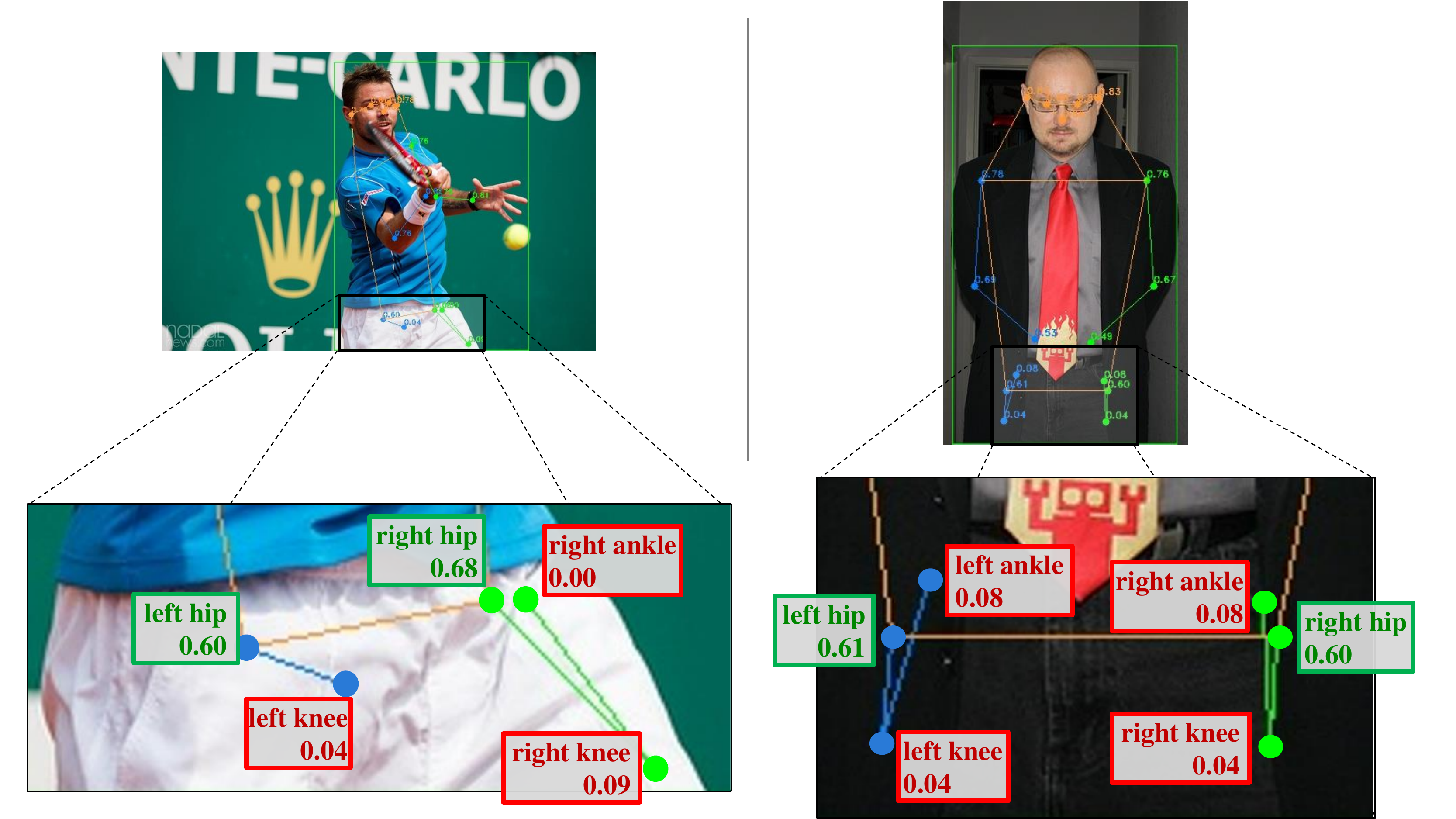}
    % }
    \vspace{-0.6cm}
    \caption{Visualization of confidence scoring. 
    There are 2 rows in each box, representing the type of joint and the confidence score prediction value, respectively. 
    The box border color represents whether the predicted joint is in the correct position (green is correct while red is wrong).
    }
    \vspace{-0.5cm}
    \label{fig:score_vis}
\end{figure}
% \vspace{-0.2cm}
\subsection{Visualization of Confidence Scoring}
% \vspace{-0.1cm}
We show that the confidence score predicted by DistilPose for 
each coordinate is plausible. 
As the examples shown in Figure~\ref{fig:score_vis}, 
most of the joint predictions are in the right position, 
and the confidence score predictions of these joints are relatively high. 
However, there are also some poor predictions($\emph{e.g.}$, the predictions denoted by red box in Figure~\ref{fig:score_vis}).
Fortunately, DistilPose can predict a low confidence score for most poor predictions while keeping high confidence in the correct predictions.
Thus we could filter these poor predictions simply by setting a threshold.
%
% 不仅在我们的任务中能提点，在下游任务和实际应用中也非常重要
A plausible confidence score not only enhances the performance of the pose estimator, but is significant in practical applications and downstream tasks for pose estimation. 
%

% \vspace{-0.3cm}
\section{Conclusion}
% \vspace{-0.1cm}
In this work, we proposed a novel human pose estimation framework, termed DistilPose, which includes Token-distilling Module (TDE) and Simulated Heatmaps to perform heatmap-to-regression knowledge distillation.  
In this way, the student regression model acquired refined heatmap knowledge at both feature and label levels, achieving a big leap in performance while maintaining efficiency.
Extensive experiments were conducted on the MSCOCO dataset to demonstrate the effectiveness of our proposed DistilPose. 
In short, DistilPose achieved state-of-the-art performance among regression-based methods with a much lower computational cost.

\section*{Acknowledgement}
This work was supported by National Key R\&D Program of China (No.2022ZD0118202), the National Science Fund for Distinguished Young Scholars (No.62025603), the National Natural Science Foundation of China (No.U21B2037, No.U22B2051, No.62176222, No.62176223, No.62176226, No.62072386, No.62072387, No.62072389, No.62002305 and No.62272401) and the Natural Science Foundation of Fujian Province of China (No.2021J01002, No.2022J06001). 
We thank Peng-Tao Jiang from Nankai University for the inspiration during our research.  

\newpage
%%%%%%%%% REFERENCES
{\small
\bibliographystyle{ieee_fullname}
\bibliography{egbib}
}

\onecolumn
\setcounter{section}{0}
\include{supplementary}
\end{document}

%% file: supplementary.tex
   \newpage
   \null
   % \iftoggle{cvprrebuttal}{\vspace*{-.3in}}{\vskip .375in}
   \begin{center}
      % smaller title font only for rebuttal
      {\Large \bf Supplementary Materials \par}
      % additional two empty lines at the end of the title
      \vspace*{24pt}
      {
      \large
      \lineskip .5em
      % \begin{tabular}[t]{c}
      %   \iftoggle{cvprfinal}{
      %     \@author
      %   }{
      %     \iftoggle{cvprrebuttal}{}{
      %       Anonymous \confName~submission\\
      %       \vspace*{1pt}\\
      %       Paper ID \cvprPaperID
      %     }
      %   }
      % \end{tabular}
      \par
      }
      % % additional small space at the end of the author name
      % \vskip .5em
      % % additional empty line at the end of the title block
      % \vspace*{12pt}
   \end{center}
\thispagestyle{empty}

\section{Comparison with latest regression SOTA Poseur}
The recent SOTA method Poseur~\cite{poseur} has FPN feature and well-designed encoder/decoder, which brings a larger computational burden while achieving higher performance.
Thus, as shown in Tabel~\ref{tab:poseur}, 1) when with the same backbone~(HRNet-W48), DistilPose achieves much higher FPS while Poseur achieves higher accuracy.
2) When with similar AP (DistilPose-S (71.6\%) \& Poseur-MobileNetV2 (71.9\%)),
the FPS of DistilPose-S is 4.73 times that of Poseur.

\renewcommand{\thetable}{A}
\begin{table}[h]
    \centering
    \small
    \setlength{\belowcaptionskip}{-0.5cm}
    % \resizebox{\columnwidth}{!}{
    \scalebox{0.75}{
    \begin{tabular}{|c|c|c|c|c|c|c|}
        \hline
        Method & Backbone & Resolution & Param & GFLOPs & mAP & FPS \\
        \hline
        \multirow{4}{*}{Poseur} & MobileNetV2 & 256$\times$192 & 11.36M & 0.5 & 71.9\% & 12.1 \\
        % \hline
         & ResNet-50 & 256$\times$192 & 33.26M & 4.6 & 75.4\% & 12.0 \\
        % \hline
         & HRNet-W32 & 256$\times$192 & 38.19M & 7.4 & 76.9\% & 5.5 \\
        %\hline
         & HRNet-W48 & 384$\times$288 & 74.27M & 33.6 & 78.8\% & 5.4\\
        \hline
        \multirow{2}{*}{DistilPose} & stemnet & 256$\times$192 & 5.36M & 2.4 & 71.6\% & 40.2 \\
         & HRNet-W48-s3 & 256$\times$192 & 21.27M & 10.3 & 74.4\% & 13.7 \\
        \hline
    \end{tabular} 
    }
    \caption{Comparison between Poseur and DistilPose on MSCOCO \emph{\textbf{val}} dataset.}
    \label{tab:poseur}
\end{table}

\section{Deviation for Basic Distribution Simulation}
We conduct an ablation study to demonstrate that predicting deviations from different directions~($\sigma_x$/$\sigma_y$ for horizontally/vertical respectively) can better help student model learn the distribution information from teacher heatmaps than predicting one deviation $\sigma$ for all directions, as shown in Tab.\ref{tab:comparison_deviation}.
We also provide a set of sample comparisons of predicted deviations based on the same input image as Fig.\ref{fig:feat-map}.
%
% The visualization of simulated heatmaps is shown in Fig\ref{fig:heatmap}.

Furthermore, we provide a set of visualization cases for comparison between teacher heatmaps and different kinds of basic distribution simulation, as shown in Fig.\ref{fig:deviation_visualization}.
The three rows in Fig.\ref{fig:deviation_visualization} from top to bottom show the local distribution of teacher heatmaps, one-deviation Basic Distribution Simulation and two-deviations Basic Distribution Simulation, respectively.
Since the deviation used for target generation during the training of teacher model is usually default to 2,
the deviations we predict are also around 2.
We can see that heatmaps generated by two-deviations are more similar to teacher heatmaps than heatmaps generated by one-deviation.

\renewcommand{\thetable}{B}
\begin{table}[h]
    \centering
    \small
    % \setlength{\abovecaptionskip}{-0.1cm}
    % \setlength{\belowcaptionskip}{-0.3cm}
    % \scalebox{0.6}{
    \resizebox{\textwidth}{!}{
    \begin{tabular}{|c|c|c|c|c|c|c|c|c|c|c|c|c|c|c|c|}
    \hline
         & mAP & Nose & Shoulder(l) & Shoulder(r) & Elbow(l) & Elbow(r) & Wrist(l) & Wrist(r) & Hip(l) & Hip(r) & Knee(l) & Knee(r) & Ankle(l) & Ankle(r)\\
        \hline 
       $\sigma$  & 71.4\% & 1.99 & 2.04 & 2.02 & 2.00 & 2.04 & 2.10 & 2.03 & 2.00 & 2.01 & 2.04 & 2.05 & 2.07 & 2.08\\
       $(\sigma_x, \sigma_y)$ & 71.6\% & (2.03, 2.02) & (2.00,1.99) & (2.01,2.04) & (2.05,2.04) & (2.02,2.05) & (2.07,2.06) & (2.02,2.07) & (1.98,2.01) & (2.01,2.05) & (1.96,2.07) & (2.00,2.07) & (2.02,2.18) & (1.93,2.14) \\
        \hline
    \end{tabular} 
    }
    \caption{Comparison between single deviation and horizontal/vertical deviations for Basic Distribution Simulation.}
    \vspace{-0.5cm}
    \label{tab:comparison_deviation}
\end{table}

\renewcommand{\thefigure}{A}
\begin{figure*}[hbp]
    \centering
    \includegraphics[width=\textwidth]{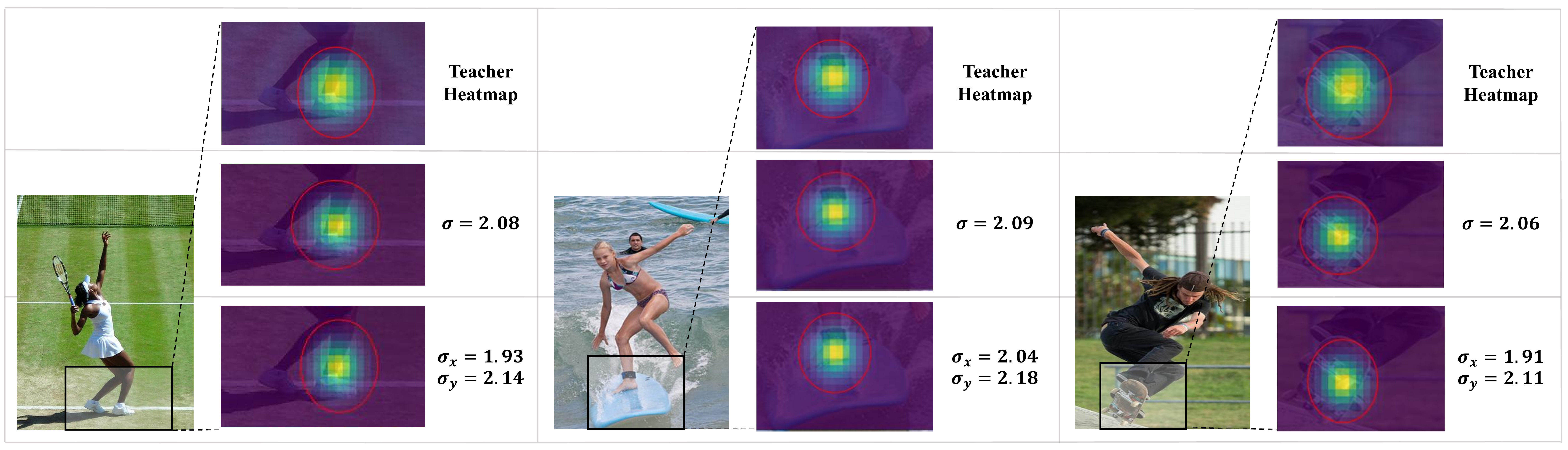}
    % }
    \caption{Visualization of teacher heatmaps and different kinds of Basic Distribution Simulation.
    In the second row, $\sigma$ represents the single deviation predicted for generating heatmaps.
    In the third row, $\sigma_x$ and $\sigma_y$ represent horizontal and vertical deviations for generating heatmaps, respectively.
    }
    \label{fig:deviation_visualization}
\end{figure*}

\section{More Visualization about Simulated Heatmaps}
We provide more visualization cases of generated heatmaps for Basic Distribution Simulation in Simulated Heatmaps, as shown in Fig.\ref{fig:supp_heatmap}.

\renewcommand{\thefigure}{B}
\begin{figure*}[hbp]
    \centering
    \includegraphics[width=\textwidth]{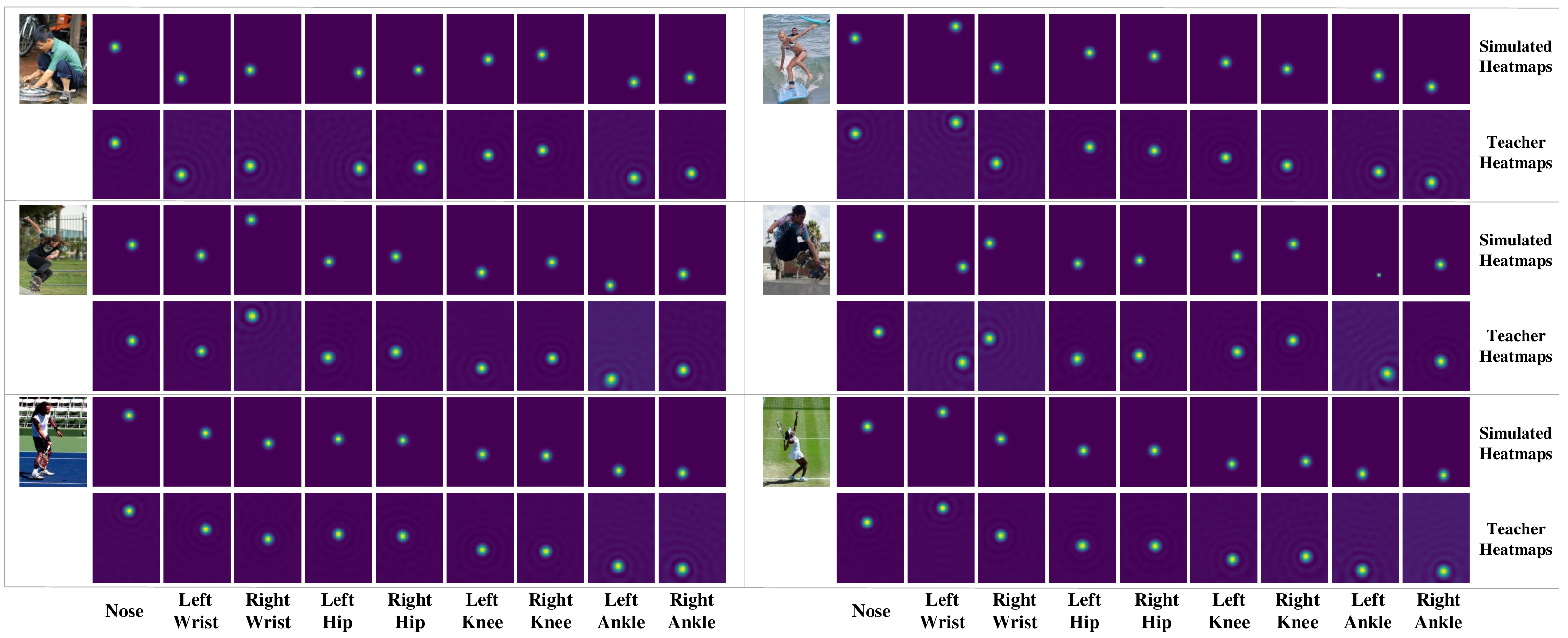}
    % }
    \caption{Visualization of Simulated Heatmaps and teacher
    heatmaps.
    }
    \label{fig:supp_heatmap}
\end{figure*}
% \newpage
% %%%%%%%%% REFERENCES
% {\small
% \bibliographystyle{ieee_fullname}
% % \bibliography{egbib}
% }

% \end{document}